\documentclass{article}
\usepackage{arxiv}
\usepackage[utf8]{inputenc}
\usepackage[T1]{fontenc}
\usepackage{hyperref}
\usepackage{url}
\usepackage{booktabs}
\usepackage{amsfonts}
\usepackage{amsmath}
\usepackage{nicefrac}
\usepackage{microtype}
\usepackage{multirow}
\usepackage{graphicx}
\usepackage{natbib}
\usepackage{doi}
\usepackage{xcolor}

\title{Capability Ceilings in Autoregressive Language Models: Empirical Evidence from Knowledge-Intensive Tasks}

\author{
  Javier Marín \\
  \texttt{javier@jmarin.info} \\
}

\begin{document}
\maketitle

\begin{abstract}
We document empirical capability ceilings in decoder-only autoregressive language models across knowledge-intensive tasks. Systematic evaluation of OPT and Pythia model families (70M-30B parameters, spanning 240× scaling) reveals that knowledge retrieval tasks show negligible accuracy improvement despite smooth loss reduction. On MMLU mathematics benchmarks, accuracy remains flat at 19-20\% (below 25\% random chance) across all scales while cross-entropy loss decreases by 31\%. In contrast, procedural tasks like arithmetic show conventional scaling where both metrics improve together. Attention intervention experiments reveal high sensitivity to perturbation: swapping attention patterns between models causes catastrophic performance collapse (complete accuracy loss) rather than graceful degradation. These measurements have immediate engineering implications: for knowledge-intensive applications using OPT and Pythia architectures, parameter scaling beyond 1-2B offers minimal accuracy gains despite continued loss improvement. Our findings quantify capability-specific scaling failures in these model families to inform resource allocation decisions. Whether these patterns reflect fundamental constraints of decoder-only architectures or implementation-specific limitations remains an open question requiring investigation across diverse architectural approaches.
\end{abstract}

\textbf{Note:} The experiments in this paper were performed in January 2024. Current model architectures are considerably more complex than those presented here.

\section{Introduction}

Organizations deploying language models face a real challenge: when does parameter scaling stop improving task-specific performance? While neural scaling laws \citep{kaplan2020scaling, hoffmann2022training} predict smooth loss improvements with model size, they do not guarantee equivalent improvement in task accuracy. This distinction becomes important for resource allocation decisions.

We provide empirical measurements documenting that autoregressive decoder-only transformers show capability-specific scaling patterns. Knowledge-intensive tasks hit accuracy ceilings despite continued loss improvement, while procedural tasks scale conventionally. 

\subsection{Empirical Contributions}

We systematically evaluated two open-source model families—OPT \citep{zhang2022opt} and Pythia \citep{biderman2023pythia}—across parameter ranges spanning 70M to 30B (240× scaling). Our measurements show the following:

\textbf{Knowledge tasks show pathological scaling:} On MMLU mathematics \citep{hendrycks2020measuring}, accuracy remains effectively constant at 19-20\% across all model sizes (below 25\% random chance for 4-choice questions), while cross-entropy loss decreases smoothly from 3.1 to 2.1. Models learn to confidently generate wrong answers.

\textbf{Procedural tasks scale normally:} Arithmetic accuracy improves from 2.4\% to 31\% as both accuracy and loss improve together, demonstrating conventional scaling behavior.

\textbf{Pattern is architecturally consistent:} The same scaling failures appear across both independently trained model families, suggesting architectural rather than training-specific limitations.

\textbf{Failures reflect learned biases:} Attention intervention experiments show strong performance degradation when manipulating learned patterns, indicating systematic biases rather than random behavior.

\subsection{Scope and Limitations}

We measure capability ceilings without explaining their mechanistic origins. Our contribution is quantifying scaling failures in widely used architectures to inform practical decisions and motivate architectural research.

\textbf{What we provide:} Measurement data showing where parameter scaling stops being useful for specific task types in decoder-only autoregressive models.

\textbf{What we don't provide:} Mechanistic explanations for why these failures occur, analysis of representation geometry, or validated theoretical frameworks.

\textbf{Generalization limits:} We evaluated two model families with similar architectural approaches. Whether these patterns generalize to other architectures (retrieval-augmented models, encoder-decoder designs, models with explicit memory) requires additional investigation.

\subsection{Practical Implications}

For knowledge-intensive applications, approaches beyond pure parameter scaling warrant investigation.
For knowledge-intensive systems deployed using decoder-only autoregressive architectures including OPT, Pythia, GPT-2, GPT-Neo, GPT-J, GPT-NeoX, or Cerebras-GPT, our measurements suggest that:

\begin{itemize}
\item Parameter scaling beyond 1-2B parameters yields minimal accuracy gains on knowledge retrieval tasks despite continued compute investment
\item Cross-entropy loss alone masks capability stagnation—accuracy measurement is essential
\item Alternative architectural approaches (retrieval-augmented generation, structured memory, continuous-space reasoning) warrant investigation for knowledge-intensive applications
\end{itemize}

Whether these patterns generalize to other decoder-only architectures (LLaMA, BLOOM, Mistral, Falcon) requires direct measurement.

\section{Related Work}

\subsection{Neural Scaling Laws}

Kaplan et al. \citep{kaplan2020scaling} established that loss scales as a power law with model size, dataset size, and compute. Hoffmann et al. \cite{hoffmann2022training} refined these relationships for compute-optimal training. However, these laws characterize loss, not task-specific accuracy. Our measurements demonstrate that loss scaling does not guarantee equivalent capability scaling across all domains.

\subsection{Emergent Capabilities and Their Critiques}

Wei et al. \citep{wei2022emergent} documented apparent "emergent capabilities" where performance jumps discontinuously at certain scales. Schaeffer et al. \citep{schaeffer2023emergent} argued these may reflect metric artifacts rather than direct transitions. Our findings suggest a third pattern: some capabilities may not emerge at all within certain architectural paradigms, regardless of scale.

\subsection{Knowledge in Language Models}

Petroni et al. \citep{petroni2019language} showed language models contain factual knowledge, while Roberts et al. \citep{roberts2020much} found this knowledge is incomplete and inconsistent. Kandpal et al. \citep{kandpal2023large} demonstrated that memorization correlates with training data frequency. Our measurements suggest that whatever knowledge these models acquire does not improve with scale in the way procedural capabilities do.

\subsection{Architectural Alternatives}

Recognition of autoregressive limitations has motivated alternative approaches. Borgeaud et al. \citep{borgeaud2022improving} demonstrated retrieval-augmented models improve factual accuracy. Guu et al. \citep{guu2020retrieval} showed similar benefits for open-domain QA. These approaches explicitly address the knowledge representation limitations we document.

Various architectural alternatives have been proposed, including joint embedding approaches \citep{lecun2022path}, though comparative evaluation across paradigms remains limited.

\section{Experimental Setup}

\subsection{Model Selection}

We selected OPT and Pythia model families based on three criteria:

\textbf{Architectural coverage:} Both represent the dominant decoder-only transformer paradigm trained with next-token prediction objectives. While not exhaustive of all possible architectures, they provide sufficient coverage to establish existence proofs of capability-specific scaling limitations.

\textbf{Scale range:} OPT provides 7 scale points (125M, 350M, 1.3B, 2.7B, 6.7B, 13B, 30B) while Pythia provides 5 points (70M, 160M, 1B, 2.8B, 6.9B), spanning three orders of magnitude.

\textbf{Documentation transparency:} Both families have publicly documented training procedures \citep{zhang2022opt, biderman2023pythia}, enabling interpretation of results in the context of training methodology.

All models were evaluated in fp16 precision using Hugging Face Transformers \citep{wolf2020transformers}.

\subsection{Task Selection}

We selected tasks covering different capability types to test scaling pattern generality:

\textbf{Knowledge-intensive (MMLU):} MMLU mathematics subtasks \citep{hendrycks2020measuring} including college mathematics, high school mathematics, and abstract algebra. We sampled 250 examples balanced across difficulty levels to ensure robust measurement.

\textbf{Procedural (Arithmetic):} Multi-digit arithmetic operations across difficulty levels (250 samples), chosen because algorithm execution should scale differently than knowledge retrieval.

\textbf{Pattern matching (QQP):} Question paraphrase detection from the Quora Question Pairs dataset (250 samples), representing surface-level pattern recognition.

\subsection{Evaluation Protocol}

For each model-task combination, we computed:

\textbf{Accuracy:} Proportion of correct answers, providing a direct measurement of capability.

\textbf{Cross-entropy loss:} $L = -\frac{1}{N}\sum_{i=1}^{N} \log p(y_i|x_i)$ measuring prediction confidence independent of accuracy.

The divergence between these metrics reveals the core phenomenon we document: models can become more confident (lower loss) while remaining equally incorrect (flat accuracy).

\subsection{Attention Intervention Experiments}

To test whether flat knowledge task performance reflects random noise versus systematic bias, we performed controlled interventions between OPT-2.7B and OPT-6.7B models. We manipulated attention patterns during inference by replacing attention weights from one model with those from another, testing three configurations:

\begin{itemize}
\item \textbf{Full replacement:} All attention layers from source model
\item \textbf{Enhanced important heads:} Weighted blend emphasizing high-variance attention heads  
\item \textbf{Constrain first half:} Replace only early layers, preserving later processing
\end{itemize}

We evaluated performance changes across knowledge-intensive (MMLU), reasoning (HellaSwag), and arithmetic tasks to assess task-specific sensitivity to attention manipulation.

\section{Results}

\subsection{Knowledge Tasks: Flat Accuracy Despite Improving Loss}

\begin{figure}[h]
\centering
\includegraphics[width=\textwidth]{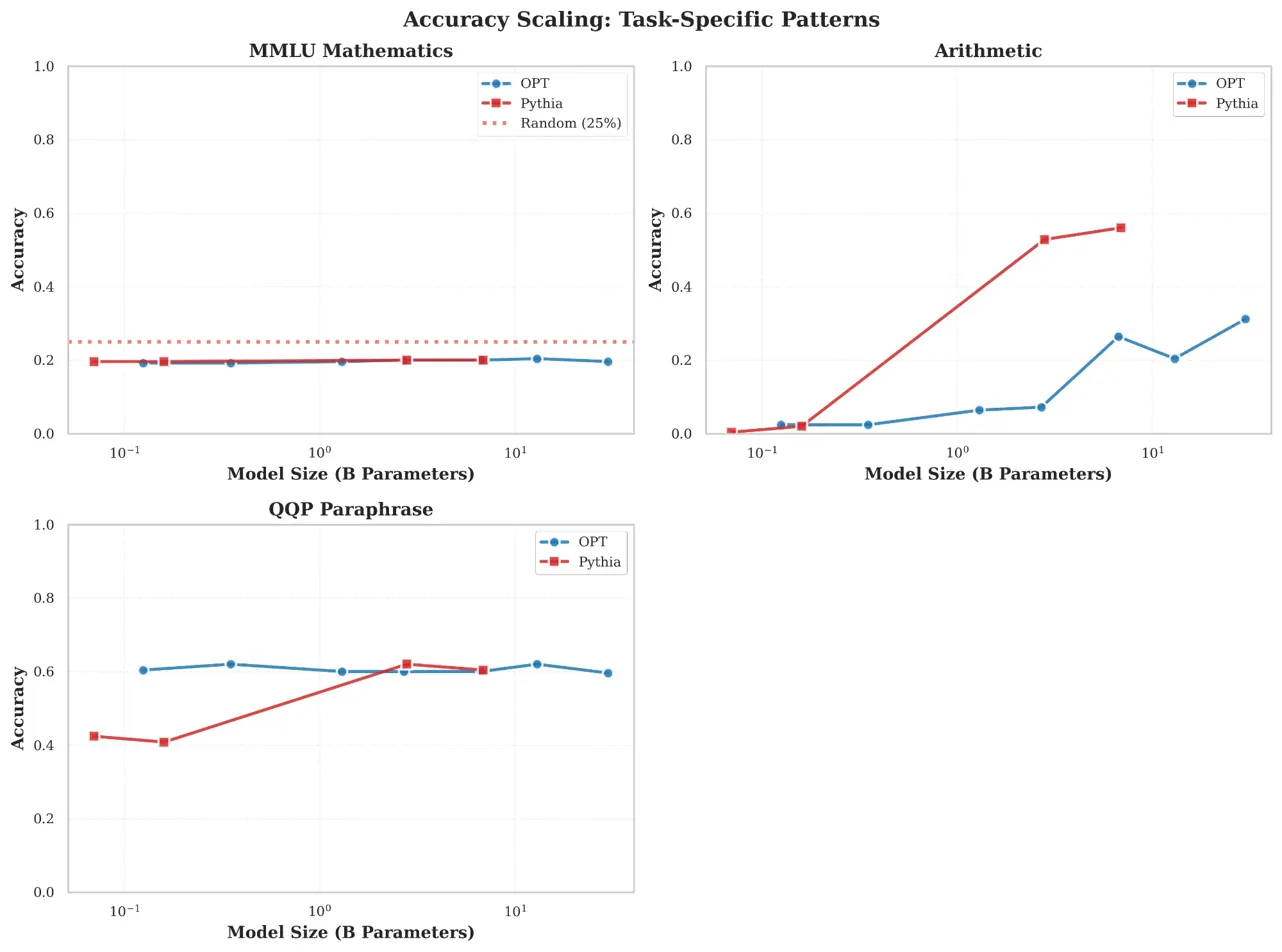}
\caption{Accuracy scaling patterns across task types. MMLU shows flat accuracy below random chance (25\%) across all scales, while arithmetic demonstrates conventional scaling. QQP (paraphrase detection) shows minimal scaling with accuracy fluctuating around 60\%. Pattern is consistent across both OPT and Pythia families.}
\label{fig:accuracy_scaling}
\end{figure}

Figure \ref{fig:accuracy_scaling} presents our primary findings across task types: MMLU mathematics accuracy remains constant at ~20\% across all scales, while arithmetic shows conventional scaling from 2\% to 31\%. QQP (paraphrase detection) shows minimal scaling with accuracy fluctuating around 60\% - suggesting this pattern-matching task saturates at small scales.

Table \ref{tab:mmlu_arithmetic} quantifies this divergence:

\begin{table}[h]
\centering
\caption{Knowledge vs. Procedural Scaling (OPT Models, 125M-30B)}
\label{tab:mmlu_arithmetic}
\begin{tabular}{lrrr}
\toprule
\textbf{Task} & \textbf{Scale} & \textbf{$\Delta$ Acc} & \textbf{$\Delta$ Loss} \\
\midrule
MMLU Math & 240× & +2.1\% & -31.1\% \\
Arithmetic & 240× & +1200\% & -26.6\%  \\
QQP & 240× & -1.3\% & -16.3\%  \\
\bottomrule
\end{tabular}
\end{table}

We observe that MMLU accuracy varies between 19.2-20.4\% across 240× parameter scaling—within measurement noise. All models perform \emph{below} 25\% random chance, indicating systematic bias toward incorrect answers. Loss decreases by 31\%, demonstrating smooth improvement in prediction confidence. Arithmetic shows 1200\% accuracy improvement with equivalent loss reduction, while paraphrase detection shows minimal scaling. In these results, we can see three distinct scaling patterns: (1) tasks where loss improves while accuracy stagnates - we term this divergent scaling, (2) tasks where both metrics improve together - coupled scaling, and (3) tasks showing limited improvement in both - saturated performance."

\subsection{Attention Interventions}

To distinguish whether flat knowledge task performance reflects robust (but ceiling-limited) knowledge versus learned statistical artifacts, we performed controlled attention interventions between OPT-2.7B and OPT-6.7B models.

Table \ref{tab:interventions} shows the results:

\begin{table}[h]
\centering
\caption{Performance Under Attention Pattern Manipulation}
\label{tab:interventions}
\begin{tabular}{llrrr}
\toprule
\textbf{Direction} & \textbf{Intervention} & \textbf{MMLU} & \textbf{HellaSwag} & \textbf{Arithmetic} \\
\midrule
\multirow{2}{*}{6.7B→2.7B} & Baseline & 0.500 & 0.500 & 0.100 \\
& Replace patterns & 0.188 (-62\%) & 0.400 (-20\%) & 0.100 (0\%) \\
\midrule
\multirow{2}{*}{2.7B→6.7B} & Baseline & 0.438 & 0.700 & 0.700 \\
& Replace patterns & 0.000 (-100\%) & 0.400 (-43\%) & 0.100 (-86\%) \\
\bottomrule
\end{tabular}
\end{table}

Attention intervention experiments reveal high sensitivity to perturbation: replacing attention patterns from differently-sized models causes severe performance degradation, with MMLU showing complete collapse (100\% accuracy loss) while arithmetic shows partial degradation (86\% loss).
This brittleness indicates that whatever these models have learned is tightly coupled to specific attention configurations. One interpretation: the models encode task-relevant information as brittle statistical patterns rather than robust, transferable representations. However, simpler explanations cannot be ruled out - the catastrophic failure may simply reflect architectural incompatibility between different-sized models rather than revealing fundamental limitations in what has been learned.
The intervention data suggests fragility but doesn't definitively distinguish between 'models learned the wrong thing' versus 'models learned in a brittle way' versus 'our intervention method is too crude to preserve learned representations across architectural changes.

\subsection{Loss Scaling Is Consistent Across All Tasks}

Figure \ref{fig:loss_scaling} shows that cross-entropy loss improves smoothly for all tasks, including those with flat accuracy:

\begin{figure}[h]
\centering
\includegraphics[width=\textwidth]{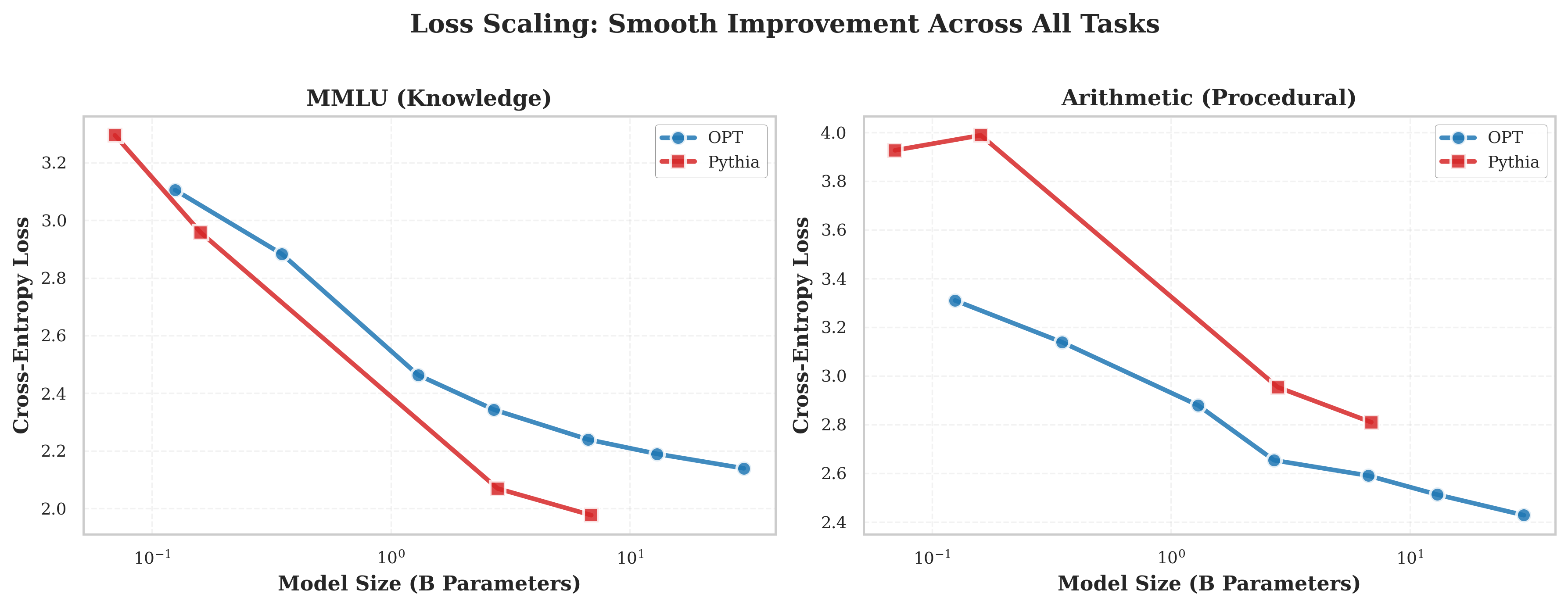}
\caption{Cross-entropy loss decreases smoothly across all tasks and both model families, regardless of whether accuracy improves. This demonstrates that loss alone masks capability stagnation.}
\label{fig:loss_scaling}
\end{figure}

Loss curves alone provide misleading signals about capability development. Models optimize their loss function (next-token prediction) without necessarily acquiring the capabilities researchers care about (factual accuracy).

\subsection{The Confidence-Competence Gap}

\begin{figure}[h]
\centering
\includegraphics[width=0.65\textwidth]{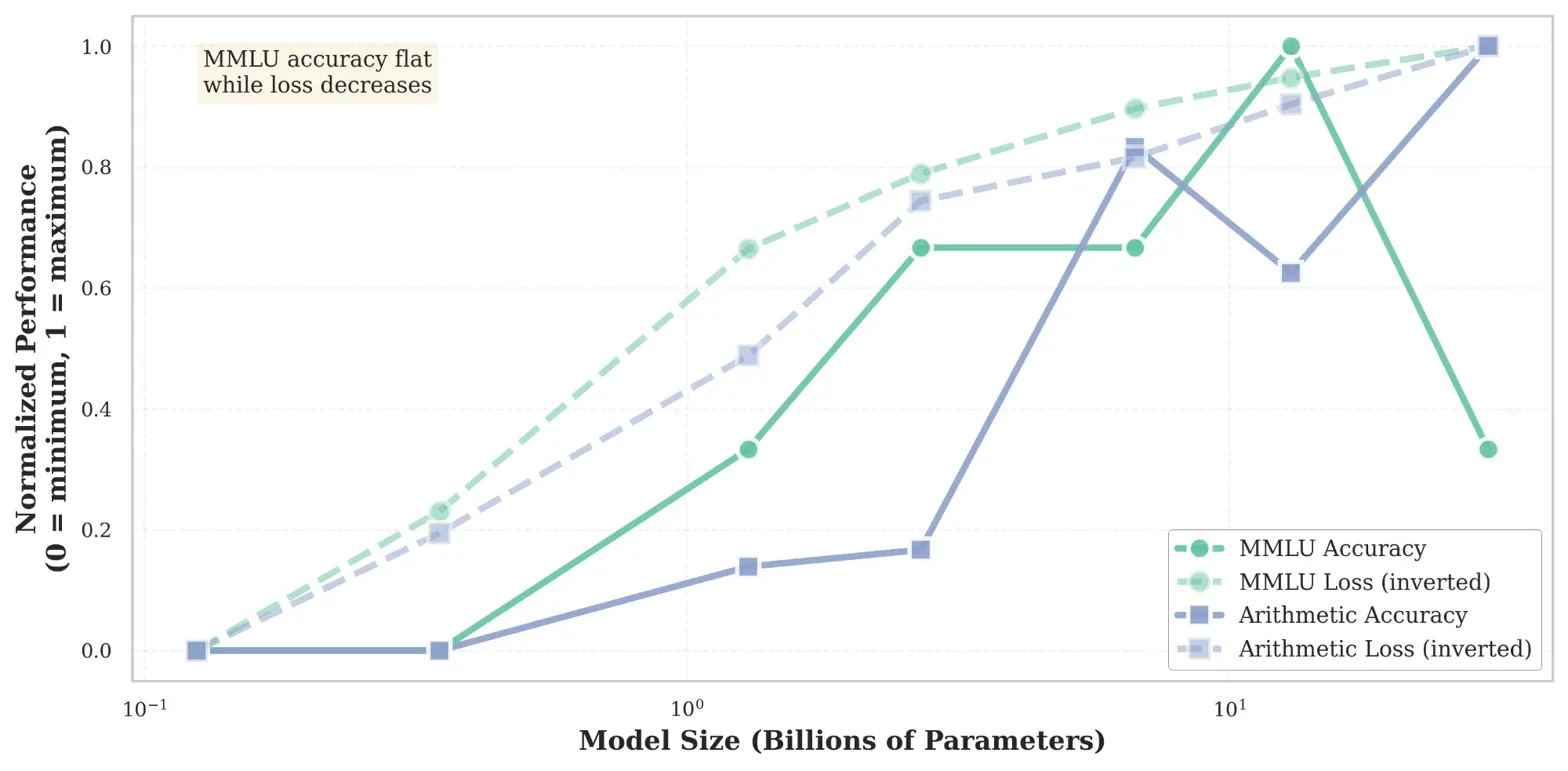}
\caption{Normalized performance metrics show MMLU loss and accuracy completely decouple as models scale, while arithmetic metrics improve together. Models learn to confidently produce wrong answers on knowledge tasks.}
\label{fig:decoupling}
\end{figure}

Figure \ref{fig:decoupling} directly visualizes the divergence between loss and accuracy. We formalize this as the \emph{confidence-competence gap ratio}:

\begin{equation}
R_{\text{CCG}} = \frac{|\Delta L| / L_0}{|\Delta A| / A_0}
\end{equation}

where $\Delta L$ is loss change, $\Delta A$ is accuracy change, and subscript 0 denotes initial values.

For MMLU: $R_{\text{CCG}} \approx 48$ (loss improves 48× faster than accuracy), and for Arithmetic: $R_{\text{CCG}} \approx 0.26$ (both improve proportionally). High ratios indicate pathological scaling where the optimization objective diverges from task performance.

\section{Discussion}

Our measurements document that MMLU mathematics accuracy remains flat at 19-20\% across OPT and Pythia models spanning 125M to 30B parameters, while cross-entropy loss decreases smoothly by 31\%. Arithmetic accuracy improves from 2.4\% to 31\% over the same scale range. This demonstrates that loss improvements do not guarantee accuracy improvements - the metrics can decouple completely.

For practitioners deploying OPT or Pythia models: parameter scaling beyond 1-2B yields minimal accuracy gains on MMLU-style knowledge tasks despite continued compute investment. Cross-entropy loss alone masks this stagnation. Task-specific accuracy measurement is essential for resource allocation decisions.

Our attention intervention experiments show that swapping attention patterns between differently-sized models causes severe performance degradation - complete collapse on MMLU versus partial degradation on arithmetic. This brittleness indicates that whatever these models encode is tightly coupled to specific architectural configurations. Whether this reflects fundamental limitations in what has been learned versus architectural incompatibility between model sizes remains unclear from our data.

We have not explained why these patterns occur. Critical questions remain unanswered: What do MMLU task representations actually look like in these models? When during training do accuracy ceilings manifest? Do these scaling failures generalize to other decoder-only architectures like LLaMA or GPT-4, or to retrieval-augmented and encoder-decoder designs? Is the MMLU ceiling specific to OPT/Pythia training approaches or inherent to the task-architecture combination?

Our work documents scaling failures in two model families without establishing whether these reflect implementation choices or fundamental constraints. Modern production systems incorporate retrieval, extensive fine-tuning, and architectural modifications we did not test. Whether those approaches address the limitations we measured remains an empirical question.

The below-random-chance MMLU performance (19-20\% on 4-choice questions) indicates systematic bias toward incorrect answers rather than random guessing. Combined with the brittleness under intervention, this suggests these specific models may have learned statistical patterns that correlate with benchmark structure rather than robust knowledge representations. However, confirming this interpretation requires mechanistic analysis of learned representations that we have not performed.

\section{Future Work}

Our measurements document scaling failures without explaining their origins. Understanding why MMLU accuracy stays flat while loss improves requires mechanistic analysis we have not performed: examining the structure of learned representations, analyzing attention patterns that our intervention experiments show are critical for performance, and tracking when during training these capability ceilings emerge.

The below-random-chance MMLU performance and brittleness under intervention suggest systematic learned biases, but we have not characterized what those biases are or how they form. Determining whether these reflect artifacts in benchmark construction, patterns in training data, or properties of the optimization process requires analysis of training dynamics and data composition that we lack access to.

Most critically, we do not know if these patterns generalize beyond OPT and Pythia. Testing whether other decoder-only architectures (LLaMA, BLOOM, Mistral), retrieval-augmented systems, or encoder-decoder models show similar scaling failures would establish whether we have documented implementation-specific limitations or broader architectural constraints.

Our work raises questions about what these models learn when optimizing next-token prediction on knowledge tasks, but answering those questions requires investigation methods we have not employed. Whether the failures we observed can be addressed through different training approaches, architectural modifications, or hybrid systems combining multiple paradigms remains empirical questions requiring systematic comparison across approaches.

\section{Conclusion}

Our measurements document that parameter scaling beyond 1-2B offers minimal MMLU accuracy gains in OPT and Pythia architectures, despite these models demonstrating strong performance on procedural tasks. Attention intervention experiments suggest knowledge task performance depends on learned statistical patterns that are brittle under perturbation.
These findings have practical implications for resource allocation when deploying these specific model families. Whether these patterns reflect fundamental constraints of decoder-only architectures or are specific to OPT/Pythia training approaches remains unclear. Modern autoregressive systems deployed in production (GPT-4, Claude, etc.) incorporate retrieval, fine-tuning, and architectural modifications we did not test.
Our work documents scaling limitations in specific implementations, not paradigm failures. The success of autoregressive models in practice suggests these limitations may be addressable through hybrid approaches, though our measurements indicate pure parameter scaling is insufficient for knowledge tasks in the architectures we tested.

\bibliographystyle{plain}

\end{document}